\pdfoutput=1
\documentclass[11pt]{article}

\usepackage{ACL2023}

\usepackage{times}
\usepackage{latexsym}

\usepackage[T1]{fontenc}
\usepackage[utf8]{inputenc}
\usepackage{microtype}
\usepackage{booktabs}
\usepackage[ruled]{algorithm2e}
\usepackage{amsmath}
\usepackage{amsfonts}
\usepackage{amssymb}
\usepackage{amsthm}
\usepackage{tabularx}
\usepackage{subfigure}
\usepackage{multirow}

\usepackage{newfloat}
\usepackage{listings}

\usepackage{array}
\newcommand{\PreserveBackslash}[1]{\let\temp=\\#1\let\\=\temp}
\newcolumntype{C}[1]{>{\PreserveBackslash\centering}p{#1}}
\newcolumntype{L}[1]{>{\PreserveBackslash\raggedright}p{#1}}

\usepackage{graphicx} % DO NOT CHANGE THIS
\usepackage{natbib}  % DO NOT CHANGE THIS AND DO NOT ADD ANY OPTIONS TO IT
\usepackage{caption} % DO NOT CHANGE THIS AND DO NOT ADD ANY OPTIONS TO IT

\usepackage{microtype}

\usepackage{inconsolata}

\title{Extrapolating Multilingual Understanding Models \\ as Multilingual Generators}

\author{
Bohong Wu\textsuperscript{\rm1}, Fei Yuan\textsuperscript{\rm2}, Hai Zhao\textsuperscript{\rm1},  
 Lei Li\textsuperscript{\rm3}, Jingjing Xu\textsuperscript{\rm2} \\
\textsuperscript{\rm 1} Department of Computer Science and Engineering, Shanghai Jiao Tong University \\
\textsuperscript{\rm 2} Shanghai Artificial Intelligence Laboratory \\
\textsuperscript{\rm 3} University of California, Santa Barbara \\
  \texttt{bohongwu@sjtu.edu.cn}, \texttt{yuanfei@pjlab.org.cn}, \\ 
   \texttt{zhaohai@cs.sjtu.edu.cn},  \texttt{lilei@cs.ucsb.edu},  \texttt{jingjingxupku.02@gmail.com}
}

\begin{document}
\maketitle

\begin{abstract}

%, and thus existing encoder-based models generally support more languages than decoder-based models
%However, with the increasing scales, the left-to-right structure brings new efficiency issues.
 
%The autoregressive framework (or decoder-based) has been a dominant choice in multilingual generative models that generates texts from left to right (e.g., GPT-3). However, despite promising results, these models bi-directional language models 
%Pre-trained multilingual understanding models (encoder-based) have shown promising results 
%Pre-trained bi-directional models have shown promising results on multilingual understanding tasks. 
%Masked language modeling has been a dominant choice in multilingual understanding models.  
%Language 
%Masked language modeling has shown promising results on multilingual understanding tasks but still struggles with handling generation tasks. With a non-autoregressive generation framework, these   
Multilingual understanding models (or encoder-based), pre-trained via masked language modeling, have achieved promising results on many language understanding tasks (e.g., mBERT). However, these non-autoregressive (NAR) models still struggle to generate high-quality texts compared with autoregressive (AR)  models.
Considering that encoder-based models have the advantage of efficient generation and self-correction abilities, this paper explores methods to empower multilingual understanding models the generation abilities to get a unified model. 
Specifically, we start from a multilingual encoder (XLM-R) and propose a \textbf{S}emantic-\textbf{G}uided \textbf{A}lignment-then-Denoising (SGA) approach to adapt an encoder to a multilingual generator with a small number of new parameters. Experiments show that the proposed approach is an effective adaption method, outperforming widely-used initialization-based methods with gains of 9.4 BLEU on machine translation, 8.1 Rouge-L on question generation, and 5.5 METEOR on story generation on XLM-R$_{large}$. On the other hand, we observe that XLM-R is still inferior to mBART in supervised settings despite better results on zero-shot settings, indicating that more exploration is required to make understanding models strong generators. %On adaption settings where the parameters of pre-trained models are fixed, XLM-R with Sgrad reaches the performance of mGPT adaptation results with XX inference speedup on tasks including translation and question generation, especially on low-resource and zero-shot settings.
%XLM-R with SGA can beat mGPT (AT pre-trained models) on translation, question generation, and story generation with notable inference speedup. 
\end{abstract}

\section{Introduction}

Multilingual encoder-based models (e.g., mBERT \cite{pires2019multilingual}, XLM-R \cite{conneau2020unsupervised}), pre-trained via masked language modeling, have demonstrated strong performance on a wide range of understanding tasks \cite{conneau2018xnli,liang2020xglue,hu2020xtreme}. Existing multilingual pre-trained models can be classified into two settings: autoregressive models \cite{liu2020multilingual,xue2021mt5,scao2022bloom} and non-autoregressive models \cite{pires2019multilingual,conneau2020unsupervised,ouyang2021ernie}. Typically, the AR framework, where a target sequence is generated from left to right, succeeds in multilingual generation tasks \cite{chen2022mtg,qi2018and}. As a comparison, encoder-based models are NAR models that are usually limited to understanding tasks \cite{conneau2018xnli}. Despite superior understanding results over AR models, these NAR models still struggle to handle a wide range of multilingual generation tasks. However, NAR models still have obvious advantages in generation efficiency and decoding flexibility \cite{gu2018non,qian2021glancing,huang2021improving,ghazvininejad2019mask,saharia2020non}, which enables  generating multiple tokens at one time in arbitrary order. Considering these strengths, this paper aims to explore methods to make multilingual encoder-based models better generators with a small number of new parameters.

\begin{figure}[t]
    \centering
    \includegraphics[width=1.0\linewidth]{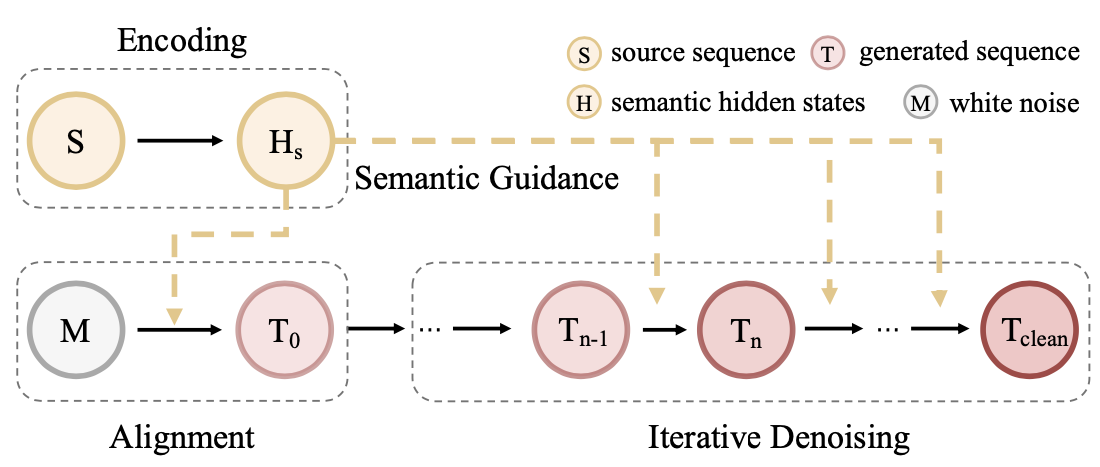}
    \caption{An overview of semantic-guided generation using pre-trained understanding models.  The encoding step is responsible for mapping the source input into a shared space that supervises the following generation.  By taking the source input and a blank sentence as input, the alignment stage generates target tokens simultaneously. Then, we feed the source representations and the generated sequence into the denoising stage for NAR denoising. The denoising step is performed iteratively until the generated text keeps unchanged or reaches the maximum loop.}
    \label{fig:info_flow}
\end{figure}

There is limited research focusing on empowering understanding models with the generation ability. Traditional methods usually use pre-trained encoders as initializers for AR models in various monolingual generation tasks \cite{su2021non}. Despite promising results, it does not satisfy our target that fixes pre-trained parameters to build a unified model for any language tasks. % but finetuning the whole pre-trained model is costly and would lose plenty of knowledge hidden in NAR pre-trained models, especially when pretrained models tend to scale up in recent years \cite{brown2020language}. 
More recently, researchers have focused on learning-free approaches~\citep{DBLP:journals/corr/abs-1902-04094,kumar2022gradient,DBLP:journals/corr/abs-2202-11705}. One typical approach is iteratively choosing tokens to mask and sampling proposals using energy models~\citep{mireshghallah2022mix}, resulting in surprising latency. Furthermore, these learning-free methods are usually limited to controllable generation and are still inferior in handling complicated tasks like machine translation. 
%sampling methods focus on empowering the English-centric model (e.g., BERT \cite{devlin2019bert}) with the ability of free generation by iteratively choosing tokens to mask and sample proposals using energy models, but sampling results in poor latency.
Unlike these monolingual studies, adapting multilingual understanding models to multilingual generation has its own challenges: semantic constraints under conditional generation where the generation process should follow the semantic constraints given source texts in any language, and parameter efficiency constraints where a single model can serve text generation in any language.

%We start from a multilingual encoder-based model. For simplification, we choose a widely used model XLM-R as an example.
We propose a semantic-guided approach to address these challenges, with a two-stage generation process: alignment-then-denoising.  The two stages share the same pre-trained parameters and only add a small number of new prefix parameters for adaptation. Given that masked language modeling (MLM) is also a denoising objective, existing multilingual pre-trained models can be naturally adapted to good denoisers. Therefore, we introduce a denoising stage into our framework. 
%Under our investigation, such adaption is technically possible since the pretraining objective MLM is considered a kind of generation task, and shares similarities with an autoregressively trained classification head. 
%First, we need 
The whole generation process is shown in Figure \ref{fig:info_flow}. 
The encoding part maps the source input into a shared space that supervises the following generation. By taking the source input and a blank sentence as input, the alignment stage generates target tokens simultaneously.
We feed the source representations and the generated sequence into the denoising module for NAR denoising. The denoising step is performed iteratively until the generated text keeps unchanged or the maximum loop is reached. %Experiments demonstrate that the denoising technique is an effective trigger to leverage the advantage of self-correlation abilities in all generations tasks.%Therefore, we propose an efficient \textbf{bi}-directional \textbf{g}eneration model BIG, with a two-stage semantic guided generation process: alignment-then-denoising. 

%First, we map source input to semantic representations. The semantic representation is then concatenated with empty representations to get a coarse-grained target sequence. Second, we feed the source representation and the generated text into the denoising stage to re-fine generation texts. Such a step is performed iteratively until the generated text remains unchanged. Experiments have demonstrated that the denoising technique is extremely helpful in all generation scenarios.
%Given the success of prompt-tuning methods in understanding tasks, MSP first uses prompt tuning to adapt a pretrained multilingual GPT model to various translation scenarios, and achieves promising results. Following MSP, we also use prompt-tuning in our adaption. 
% Specifically, we combine prompt-tuning with existing NAR methods, including CTC and Mask-Predict as our learning objective, and validate our model on various generation tasks including multilingual translation and multilingual generation. 

%For parameter-efficient tuning techniques, following MSP \cite{tan2022msp}, a multi-stage prompt-tuning-based method that efficiently adapts a multilingual GPT model to translators, we also use prompt tuning for efficient adaptation. 

Experiments demonstrate that our model has achieved better results on various generation tasks than traditional fine-tuning-based approaches that directly use NAR pre-trained models as initialization, with gains of 9.4 BLEU on machine translation, 8.1 Rouge-L on question generation, and 5.5 METEOR on story generation on XLM-R$_{large}$. More promisingly, our method has achieved impressive zero-shot cross-lingual ability in translation tasks, outperforming a multilingual AR adaptation model, mGPT + MSP by a large margin. %Such cross-lingual ability further enables us with the possibility to share prompts across all language directions. On machine translation tasks, we only need as few as 21M parameters to adapt a pre-trained multilingual understanding model to an effective unified translator. 
On the other hand, we also notice the gap between XLM-R and AR models. Generally, XLM-R with fine-tuning is largely inferior to mBART fine-tuning. With our methods, the gap is largely reduced but still exists. In future work, we would like to explore pre-training methods to make multilingual understanding models  better generators.
% The whole generation, which is shown in Figure \ref{fig:info_flow} contains two stages: alignment-then-denoising. 

% The whole denoising process adopts MCMC sampling during training. We keep running this step until we reach the maximum loop or we reach the steady state in MCMC. 
Our contributions can be summarized as follows:

\begin{itemize}
    \item We propose an efficient adaptation method to make multilingual understanding models better generators in a parameter-efficient way. 
    
    \item We present a semantic-guided denoiser, which can efficiently improve the generation quality. 
    
    \item Experiments show that our proposed method outperforms traditional initialization-based adaptation methods by a large margin. 
\end{itemize}

% We also propose a new training recipe to train such a semantic-guided MCMC generation framework. The CTC loss is adopted during alignment and denoising with true target output as supervision. To train a better build The denoising largely affects the performance. We regarded the generation  

%With the current need for plug-and-play controllable text generation, causal models, including encoder-decoder architectures (mBART, mT5) and decoder-only architectures (GPT etc.), are often preferred, as we can easily control the generation behavior of these models by greedy sampling or nucleus sampling. Bidirectional understanding models, on the contrary, are often regarded as lacking of ability for high-quality generation and controllable text generation due to their inductive bias. 

%\clearpage

\begin{figure*}
    \centering
    \includegraphics[width=0.95\linewidth]{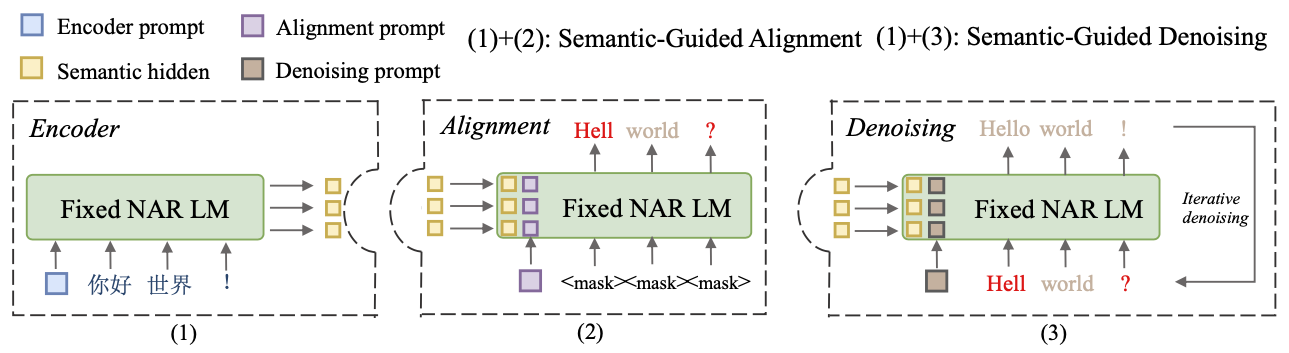}
    \caption{An overview of the generation units. All units share the same pre-trained parameters and individual prompt parameters (shown in blue square, purple square, and gray square).  The encoder maps the source input into a sequence of hidden representations (yellow), which are then fed into a decoder and a denoiser for target generation. The alignment unit is responsible for generating a piece of target text. The denoiser is responsible for refining the generated text.}
    \label{fig:main_arch}
\end{figure*}

\section{Related Work}
In this section, we review the related studies including parameter-efficient adaptation, adapting encoder models as generators, and non-autoregressive generation. 
\paragraph{Parameter efficient adaptation}
% 几个可能要展现出来的点？
% 1. 所有parameter efficient的方案，最后在说prompting的作用 -- mixed task inference -- 再引一些应用上的论文，最后说MSP的时间效率等等的问题

% 两个点，一个是model is capable of these tasks
% 第二个点就是怎么做好
Pre-trained language models (PLMs) \cite{devlin2019bert,liu2019roberta,clark2019electra,conneau2020unsupervised} have achieved overwhelming performance in a variety of downstream tasks. %For consideration of low disk requirement and dealing with task-interference in multi-task and multilingual scenarios, 
Parameter-efficient tuning is a hot research direction to adapt PLMs to downstream tasks with only a few parameters. Adapter-based methods \cite{bapna2019simple} are one of tge popular parameter-efficient approaches. Recent studies~\citep{ustun2021multilingual,stickland2021recipes} proposed to use adapters on the top of an mBART \cite{liu2020multilingual} model, enabling a flexible and well-performed method for plug-and-play translation. More recently, prompting-based methods~\citep{li2021prefix,lester2021power,liu2022p,tan2022msp} have proved to be extremely helpful, and can easily support mixed-task inference as it does not require changing the architecture of PLMs. In this work, we follow this research thread for efficient adaptation and apply prompt-based approaches in our work. % \citet{li2021prefix} proposed prefix tuning for table-to-text generation and summarization. 
%\citet{liu2022p} proposes P-Tuning and achieve comparable performance with fine-tuning across a variety of understanding tasks. 
%\citet{tan2022msp} further proposed multi-stage prompting method for multilingual translation, based on a multilingual GPT model. However, it requires multiple passes through one large language model, making the time latency prohibitively unbearable.

% Few researches (except for MSP \cite{tan2022msp}) explore prompting methods on generation tasks, as these tasks need a well pre-trained decoding space, which usually requires a lot of parameters to fine-tune, contradicting with the requirement of parameter efficient fine-tuning. In this paper, we show that simply Masked Language Modelling shares much similarity with translation tasks, and by simple prompting can an existing PLM be adapted to multilingual translation.

% Few studies explore whether bi-directional pre-trained models can be applied to translation or generation tasks, as there exists obvious inductive bias. However, we found it somehow possible.

\paragraph{Adapting encoder-based models for generation}
%Because of good zero-shot cross-lingual ability in existing multilingual encoder-based models~\cite{conneau2020unsupervised}, 
Previous works proposed to use pre-trained understanding models to initialize encoder-decoder models~\citep{chen2021zero,ma2021deltalm}. With the trend of scaling up models in recent years, it gradually becomes impossible to fine-tune the whole language model in each language direction. %Moreover, initializing an AR transformer model using XLM-R weights often leads to inferior performance, unless sophisticated fine-tuning strategies are adopted \cite{chen2021zero}.
In addition, there have been several learning-free methods that adopt encoder models as energy scorers for controllable text generation \cite{mireshghallah2022mix,kumar2022constrained}. Although these methods do not need to fine-tune the pre-trained model, they require multi-steps of sampling and refinement, resulting in surprising inference latency. %Despite promising, these researches are limited to monolingual generation scenarios and require multiple steps of sampling, which totally loses the advantage of generation speed for non-autoregressive models.

% Mix\&Match \cite{mireshghallah2022mix} propose to use Gibbs sampler with Metropolis-Hastings correction to iteratively revise the given text to satisfy given constraints. MUCOLA \cite{kumar2022constrained} uses a gradient-based method and iteratively refines the input via langevin dynamics on scenarios controllable text generation. 

% With the scaling of multilingual models, how to efficiently adapt existing encoder models to generation tasks remains challenging and worth exploring. Though many works have studied how to efficiently adapt pre-trained autoregressive models to all generation tasks, autoregressive models are in some aspects inferior to non-autoregressive models due to left-to-right decoding strategy and time latency. This work then explores how to efficiently adapt existing encoder models to various generation tasks.

\paragraph{Non-autoregressive generation}
Our work aims at adapting multilingual encoders to multilingual generators, instead of developing a NAR architecture like previous NAR literature does. Therefore, there is a lot of difference between our work with previous NAR studies. Despite different motivations, our implementation also uses several NAR techniques, like CTC~\cite{libovicky2018end} and Mask-Predict~\cite{ghazvininejad2019mask}.  For clarification, we also review the thread of NAR generation.  %Despite the impressive performance for autoregressive models in sequence-to-sequence modeling, non-autoregressive generation methods \cite{gu2018non} are still attracting the attention of researchers for their latency. 
Single-step NAR generation is a popular research direction that generates text at one time. To mitigate the gap between single-step NAR methods and AR methods, researchers have proposed alignment-based methods \cite{libovicky2018end,ghazvininejad2020aligned, du2021order} or glancing-based methods \cite{qian2021glancing}. 
%However, the performance still lags far behind AR methods, and often requires knowledge distillation \cite{gu2018non} to reduce data modalities.
As a compromise, iterative NAR methods can provide both comparable performance and better latency with AR baselines~\cite{lee2018deterministic,ghazvininejad2019mask,huang2021improving,saharia2020non}. For example, SUNDAE~\cite{savinov2021step} proposed step-unrolled denoising, and achieved good performance in both machine translation and text infilling. %Iterative methods  propose to uncover the target sequence step by step based on their confidence score, 
Similar iterative idea has been adopted at recent diffusion models~\citep{li2022diffusion,gong2022diffuseq}.
In this work, we adopt an iterative decoding idea to take advantage of the denoising abilities of encoder-based models which are pre-trained with denoising objectives.

% There are also many researches studying the pretraining on monolingual NAT \cite{qi2021bang,li2022elmer}, however, very limited research study NAT pertaining in the multilingual setting.
%Recently, researchers have been exploring denoising-based methods in non-autoregressive sequence generation. The research area is not restricted to translation only, too.
% Recent development of diffusion models , have also shown overwhelming performance in generation tasks other than translation, including controllable text generation and question generation tasks.

\section{Notation and Background}

\paragraph{Prompt tuning} For efficient adaption, we follow mGPT+MSP \cite{tan2022msp} and use prompt tuning to adapt an existing pre-trained NAR pre-trained multilingual understanding models to generators. Formally, we denote $K^{l}$ and $V^{l}$ as the key-value pairs in the $l$-{th} Transformer layer. The introduced prompt-tuning parameters are $(K,V)$ pairs, which will be concatenated with current key-value pairs during training and inference. In prompt tuning, we denote the forward pass of a pre-trained LM as $f_{LM}(\theta_{p}, X)$, which accepts two inputs including prompt parameters $\theta_{p}$ and the source sequence $X$. %Our goal is to generate the target sequence $Y$.

\section{SGA Approach}

% We propose a two-stage while parameter efficient method to adapt one understanding model to multiple generation scenarios. Although there are various kinds of PEFT methods including adapters and prefix-tuning, following MSP \cite{tan2022msp}, we use prompt tuning, as it not only achieves comparable performance, but also endues large language models with mixed-task inference ability. We directly use the off-the-shelf pre-trained multilingual understanding LM, XLM-R \cite{conneau2020unsupervised}.

\subsection{Overview}

Based on a fixed multilingual understanding model, Figure \ref{fig:main_arch} presents the overview of our proposed SGA, which contains three stages, including semantic encoding, alignment and then denoising.  Section \ref{method:encode} presents the semantic encoding unit, which maps sentences in all languages to a unified space. Section \ref{method:align} presents the alignment unit, which generates a target sentence. Section \ref{method:align} presents the denoising unit, which refines the generated sentences under the guidance of semantics.

\subsection{Semantic Encoding} \label{method:encode}

% We adopt the encoder-decoder architecture as our non-autoregressive generator. For both parameter efficiency and fully explore the generation ability of understanding models under mixed-task inference scenarios, we remove the cross-attention structure.

%Considering that generation is a process that recovers target sentences from white noise under the guidance of semantic meaning, we expect an understanding model can extract the semantic meaning of the source sequence. 

% More importantly, we expect a multilingual model can parameter-efficiently project sentences of different languages into a unified embedding space, which can effectively utilize the cross-lingual ability of pre-trained multilingual LMs.

% Such semantic meaning could be utilized in the latter alignment and denoising stage to provide guidance. 

Suppose a pre-trained multilingual language model with $L$ layers. we define prompt parameters $\theta_{p_S}= (K_{S}^{1:L}, V_{S}^{1:L})$ for all layers. These parameters are then concatenated with the key-value  pairs of attention in each layer to extract the hidden representation of the source sequence $X=[x_1, x_2, ..., x_t]$. Therefore, we can get the layer-wise hidden representation $h_{S}^{1:L}$ via
\begin{equation}
    h_{S}^{1:L} = f_{LM}(\theta_{p_S}, X)
\end{equation}

For semantic guidance, we directly pass $h{S}^{1:L}$ to the alignment unit and denoising unit as new prompt parameters. We use additional two projection layers $W_K$ and $W_V$ to project $h_{S}^{1:L}$ into semantic hidden states, denoted as $K_{S}^{1:L}$ and $V_{S}^{1:L}$.
\begin{equation}
\begin{split}
    K_{S}^{1:L} = h_{S}^{1:L} W_K\\ V_{S}^{1:L} = h_{S}^{1:L} W_V
\end{split}
\end{equation}

\subsection{Semantic-guided Alignment} \label{method:align}

This unit generates the target sequence in parallel, by taking a sequence of white noise as input, denoted as $Y_{\text{blank}}$ (a sequence of \textit{<mask>} tokens in our experiments). Similarly, we introduce alignment prompt $\theta_{p_A}= (K_{A}^{1:L}, V_{A}^{1:L})$ to efficiently adapt the pre-trained model to generate a target sequence. To grab information from the source sequence, we directly concatenate $(K_{A}^{1:L}, V_{A}^{1:L})$ and $(K_{S}^{1:L}, V_{S}^{1:L})$, where we get 
\begin{equation}
\begin{split}
    \theta_{p_A} = (\text{concat}([K_{S}^{1:L}, K_{A}^{1:L}]),\\ \text{concat}([V_{S}^{1:L}, V_{A}^{1:L}]))
\end{split}
\end{equation}

The alignment output is obtained via:
\begin{equation}
   T_{0} = f_{LM}(\theta_{p_A}, Y_{\text{blank}})
\end{equation}

% The objective of the alignment stage can be written as
% \begin{equation}
%     \mathcal{L}_{1} = -\mathbb{E}[\log p_{(\theta_{p_S},\theta_{p_A})}(Y|X)] \\
% \end{equation}
% where $X, Y$ are the source and target sequence, $p$ is the probability function.

Formally, we define the alignment loss as $\mathcal{L}_1(\theta_{p_S}, \theta_{p_A})$ given training pair $(X, Y)$ sampling from a dataset. For alignment, we use two variants of non-autoregressive loss to train new parameters, including Connectionist Temporal Classification(CTC) \cite{libovicky2018end} and Mask-Predict \cite{ghazvininejad2019mask}. For constrained generation tasks, specifically, translation, we choose CTC loss objective for its efficiency in speed, as it is a one-step NAR generation method. For free generation tasks, CTC loss performs poorly because free-generation tasks intensify the multi-modality problem \cite{gu2018non}, which we will discuss in Appendix \ref{sec:tradeoff}. On the contrary, iterative methods choose the best possible modality during early iterations of generation. Therefore, we use Mask-Predict, which is an iterative NAR generation method that sacrifices speed for performance. 

%which will introduce a very small amount of extra parameters for length prediction. We follow the design of \citep{ghazvininejad2019mask} for length prediction. As an iterative method, Mask-Predict style alignment sacrifices the inference speed for better generation quality, but still outperforms mGPT+MSP, as mGPT+MSP requires multiple passes through a pre-trained mGPT model for the generation of each token.

% Free generation tasks introduces much more possibilities than translation, which makes single-step NAR technique inapplicable as it intensify the multi-modality problem. As an iterative method, it sacrifices the inference speed, but our proposed method still outperforms MSP in speed, as MSP requires multiple passes through a pre-trained mGPT model for the generation of each token.

\subsection{Semantic-guided Denoising} \label{method:denoise}

Due to the limitation of trainable parameters and the non-autoregressive nature, the generation result of the first-stage alignment is usually far from satisfying. Thanks to the denoising pre-training objective MLM, current language models can be easily adapted to a denoiser. In this step, we also add prompt parameters for denoising $\theta_{p_D} = (K_{D}^{1:L}, V_{D}^{1:L})$ to efficiently adapt the understanding model to a language-specific denoiser. Similarly, we get semantic-guided denoising prompt by the following equation:
\begin{equation}
\begin{split}
    \theta_{p_D} = (\text{concat}([K_{S}^{1:L}, K_{D}^{1:L}]),\\ \text{concat}([V_{S}^{1:L}, V_{D}^{1:L}]))
\end{split}
\end{equation}

We take the output sequence in alignment stage as input which is denoted as $T_0$. To avoid overfitting, we add random noise including random deletion or repetition to sequence $\Tilde{T}_0 = T_0 + \epsilon$. We can then acquire the denoised logits $T_{1}$ by:
\begin{equation}
   T_1 = f_{LM}(\theta_{p_D},\Tilde{T}_0)
\end{equation}
We repeat this step and treat $T_1$ as new input to get $T_2$. The loop is running until the output sequence keeps unchanged or we reach the maximum loop number.

%Formally, We write the denoising loss $\mathcal{L}(\theta_{p_S}, \theta_{p_D}$

% in the second stage as
% \begin{equation}
% \begin{split}
%     \mathcal{L}_{2} = -\mathbb{E}_{\Tilde{Y}^{\prime} \sim Y^{\prime} + \epsilon
%     }[\log f_{(\theta_{p_S}, \theta_{p_D})}(Y|\Tilde{Y}^{\prime})]
% \end{split}
% \end{equation}

%As illustrated in Figure \ref{fig:info_flow}, the denoising process will be iteratively adopted for better generation quality. 
For denoising, we use a CTC-based denoiser after the alignment process and adopt the CTC loss $\mathcal{L}_2(\theta_{p_S}, \theta_{p_D})$ given training pair $(Y, T_i)$ where $T_i$ is the output sequence at the i-th step. For translation, the outputs of the alignment stage are directly fed to the denoiser. For other generation tasks, we upsample the alignment result by a factor of 2 by duplicating each token, and then fed the duplicated sequence to the CTC-based denoiser.

\subsection{Training Objective}
The final loss is a combination of the alignment loss and denoising loss by the following equation:
\begin{equation}
\begin{split}
    \mathcal{L} = \mathcal{L}_{1}(\theta_{p_S}, \theta_{p_A}) 
+ \mathcal{L}_{2}(\theta_{p_S}, \theta_{p_D})  
\end{split}  
\end{equation}

% Still we could compare the speed with MSP.

% Note that we choose CTC as our training objective because it naturally relaxes the constraints of fixed position embeddings by up-sampling the sequence length. As we shown in Figure \ref{fig:main_arch}, under the non-autoregressive generation framework, it requires target sentence reconstruction from a totally masked sequence. These \textit{<mask>} tokens share very similar embeddings, as they only differ in the positional embeddings. The similarity is especially high in the neighboring tokens, which severely increases the repetition problem in NAR generation. 

% By CTC objective, the loss is written as
% \begin{A}
%     L_{CTC} & = \log P(Y\mid X)\notag \\
%             & = \log \mathbb{E}_{z\mid X}[P(Y\mid z, X)],z \in A
% \end{A}
% where $A$ is all the CTC alignments that can be mapped to $Y$.

% By Mask-Predict objective, the loss objective is the average MLE loss of all masked tokens:
% \begin{A}
%     L_{CTC} & = \log P(Y\mid X)\notag \\
%             & = \log \mathbb{E}_{z\mid X}[P(Y\mid z, X)],z \in A
% \end{A}

\section{Experiments}
%Since we share the vocabulary with XLM-R for all baselines, the parameters of bilingual models is larger than previous works, where word embedding occupies a large number of parameters

\subsection{Settings} \label{sec:settings}

% Also, we use WMT14 English-German dataset (4.5M) and WMT16 Romanian-English dataset (0.6M) to present the bilingual translation performance. We directly used the public available distilled version of WMT14 En-De and WMT16 Ro-En.
%\noindent\textbf{Datasets}
We run experiments on three multilingual generation datasets including machine translation, question generation, and story generation. Mapping between language codes and full names of all languages used in our paper is presented in Appendix \ref{append:lang_code}.

\paragraph{Dataset} For experiments on both bilingual and multilingual translation, we use TED dataset \cite{qi2018and}. We focus on English-centric settings and choose 10 languages (Ar, De, Es, Fr, He, It, Ro, Ru, Tr, Vi) with the most training data to construct our multilingual translation task. We choose five additional languages (Kk, Be, Eu, Ms, Bs) with the least training data (less than 6k) for zero-shot cross-lingual evaluation. Details are presented in Appendix \ref{append:ted_detail}.

% XGLUE-QG covers 6 languages including En, Fr, De, Es, It and Pt, with a train/dev/test split of 100000/10000/1000.
% MTG-SG collects parallel data from 5 languages including En, De, Fr, Es, and Zh, with a train/dev/test split of 15000/2000/3000 for each language pair.
For experiments on question generation, we use the Question Generation (QG) split of XGLUE dataset~\cite{liang2020xglue}. Since XGLUE only provides a training set in the English-English direction, we use M2M-100-418M \cite{fan2021beyond} to translate the English training set to all other languages. We train all models in the En$\rightarrow$X directions and evaluate them on the X$\rightarrow$X test sets. For simplicity, we report results on En$\rightarrow$En, En$\rightarrow$De and En$\rightarrow$Fr, where results on En$\rightarrow$En represent the monolingual generation ability, and results on En$\rightarrow$De, En$\rightarrow$Fr represents zero-shot cross-lingual generation ability.
For experiments on story generation, we use the Story Generation (SG) split of MTG dataset \cite{chen2022mtg}. For simplicity, we report monolingual generation result on En$\rightarrow$En, and cross-lingual generation results on De$\rightarrow$En and Fr$\rightarrow$En.

\paragraph{Implementations} \label{append:hyperparam}

We use a batch size of 32k to train all transformer models in both AT and NAT. Following \citep{xu2021vocabulary}, we use the transformer-big setting with a learning rate of 5e-4 and a dropout rate of 0.3. We train these models for a maximum of 50 epochs, and average the 5 best checkpoints for inference. We use Fairseq~\cite{ott2019fairseq} for implementation.

For fine-tuning using pre-trained language models like XLM-R and mBART, we use a batch size of 4k tokens and a much smaller learning rate of 3e-5. We train the pre-trained models for a maximum of 80,000 steps. We also use Fairseq.

For adaptation methods on PLMs, we directly follow the hyperparameter setting of~\citep{tan2022msp}, with a batch size of 32k tokens and a learning rate of 7e-4. We train these models for a maximum of 40,000 steps, and average the 5 best checkpoints for inference. We use THUMT~\cite{tan2020thumt} for the implementation of the adaptation methods. For translation tasks, the training takes around 40 hours on 8 GPUs in each translation direction to adapt an XLM-R$_{large}$ model to generators.

\paragraph{Evaluation Metrics} We calculate case-sensitive \textbf{BLEU} \cite{papineni2002bleu} using the sacrebleu toolkit \cite{post2018call} for translation evaluation \footnote{Signature:nrefs:1|case:mixed|eff:no|tok:13a|smooth:exp|
version:2.0.0}. We use \textbf{ROUGE-L} \cite{lin2004rouge} for both question generation and story generation. We also report 
\textbf{METEOR} \cite{banerjee2005meteor} for story generation. For speed calculation, we average the running time on the test set with batch size set to 1.
%, since we find using transformer-big leads to overfiting on En$\rightarrow$X translation scenarios

\begin{table*}[htp]
    \centering
    % \belowrulesep=-0.5pt
    % \aboverulesep=0pt
    % \small
    % \setlength{\tabcolsep}{2.5pt}
    \footnotesize
    \resizebox{0.97\textwidth}{!}{
    % \fontsize{10.0pt}{\baselineskip}\selectfont
    \begin{tabular}{l|lrrrrrrrrrrrrc}
         \toprule
         \textbf{Group} & \textbf{Model} & \textbf{Param.} & \textbf{Speed} & \textbf{Ar$\rightarrow$En} & \textbf{De$\rightarrow$En} & \textbf{Es$\rightarrow$En} & \textbf{Fr$\rightarrow$En} & \textbf{He$\rightarrow$En} & \textbf{It$\rightarrow$En} 
 & \textbf{Ro$\rightarrow$En} & \textbf{Ru$\rightarrow$En} & \textbf{Tr$\rightarrow$En} & \textbf{Vi$\rightarrow$En} & \textbf{Avg.} \\
         \midrule
         % \multicolumn{14}{l}{\textit{Bi-lingual Models}} \\
         % AT & 31.45 & 36.36 & 42.59 & 40.72 & 37.49 & 38.79 & 35.60 & 24.95 & 26.75 & 27.48 & 1$\times$ \\
         \multirow{2}{*}{\textit{Bi-lingual}} & Transformer (AT) & 432M & 1.0$\times$ & 32.1 & 36.0 & 41.9 & 40.5 & 38.1 & 38.4 & 35.5 & 24.7 & 26.1 & 27.1 & 34.0 \\
         % Transformer (NAT) & 173M & 23.8$\times$ & 5.9 & 5.6 & 9.0 & 8.6 & 8.9 & 7.1 & 6.0 & 4.0 & 2.3 & 5.3 & 6.3 \\
         & Transformer (NAT) & 434M & 13.4$\times$ & 17.6 & 16.2 & 29.0 & 26.1 & 23.3 & 24.0 & 19.4 & 7.1 & 0.7 & 12.7 & 17.6 \\
         
         % \midrule
         % \multicolumn{14}{l}{\textit{Multilingual Models}} \\
         \midrule
         \multirow{2}{*}{\textit{Multi-lingual}} & mTransformer & - & 0.8$\times$ & 22.2  & 27.9 & 34.5 & 32.7 & 25.8 & 30.5 & 27.5 & 19.9 & 17.6 & 20.8 & 25.9 \\
         & \quad+adapter & 50M & 0.8$\times$ & 28.0 & 32.4 & 38.3 & 36.5 & 33.2 & 34.7 & 31.8 & 21.8 & 22.3 & 24.1 & 30.3 \\
         % \midrule
         % \multicolumn{14}{l}{\textit{Adapt PLM}} \\
         \midrule
         % mBART-fine-tune & 610M & & 35.2 & 41.0 & 46.3 & 44.4 & 40.5 & 42.8 & 40.4 & 29.0 & 31.0 & 31.3 & \\
         % NAT (606M) & & & & & & & & & & & \\
         %mTransformer \\
         \multirow{9}{*}{\textit{PLM Adaptation}} & mGPT + MSP\footnotemark[1] (AT) & 19M & 0.2$\times$ & 26.2 & 29.8 & 38.9 & 36.2 & 30.3 & 33.1 & 30.9 & 21.9 & 19.4 & 23.3 & 29.0 \\
         & \multicolumn{14}{l}{\textit{XLM-R$_{base}$}} \\
         & \quad + AT initialization & 390M & 0.9$\times$ & 16.6 & 23.5 & 29.5 & 26.7 & 21.0 & 24.9 & 22.7 & 16.1 & 17.6 & 15.7 & 21.4 \\

         & \quad + SGA w/o. denoising & 6M & 6.8$\times$ & 24.6 & 29.9 & 36.1 & 32.6 & 29.7 & 30.4 & 28.7 & 18.0 & 15.1 & 23.5 & 26.9 \\
         & \quad + SGA & 8M & 3.0$\times$ & 27.1 & 33.0 & 40.0 & 35.5 & 33.3 & 32.8 & 30.3 & 20.3 & 19.6 & 23.9 & 29.6 \\
         % Ours$_{base}$-zero-shot & 7.25 & - & 14.38 & 14.59 & 10.11 & 10.57 & 8.03 & 11.78 & 3.15 & 9.71 \\
         & \multicolumn{14}{l}{\textit{XLM-R$_{large}$}} \\
         & \quad + AT initialization & 960M & 0.6$\times$ & 19.2 & 25.7 & 32.4 & 29.9 & 23.4 & 28.4 & 24.9 & 21.5 & 17.4 & 18.3 & 24.1 \\
         & \quad + SGA w/o. denoising & 15M & 3.7$\times$ & 28.2 & 33.8 & 37.9 & 36.1 & 35.5 & 34.1 & 31.5 & 21.4 & 23.4 & 24.4 & 30.6 \\
         & \quad + SGA & 21M & 1.9$\times$ & \textbf{30.7} & \textbf{37.0} & \textbf{40.9} & \textbf{38.6} & \textbf{38.5} & \textbf{37.5} & \textbf{34.2} & \textbf{24.0} & \textbf{27.2} & \textbf{26.4} & \textbf{33.5} \\
         % \quad + prompt sharing & \textbf{21M} & 4.2$\times$ & 29.1 & 34.2 & 39.3 & 38.3 & 34.0 & 36.2 & \textbf{34.5} & 23.8 & 22.7 & 26.2 & 31.8 \\
         \midrule
         \midrule

        Group & \textbf{Model} & \textbf{Param.} & \textbf{Speed} & \textbf{En$\rightarrow$Ar} & \textbf{En$\rightarrow$De} &\textbf{En$\rightarrow$Es} & \textbf{En$\rightarrow$Fr} & \textbf{En$\rightarrow$He} & \textbf{En$\rightarrow$It} & \textbf{En$\rightarrow$Ro} & \textbf{En$\rightarrow$Ru} & \textbf{En$\rightarrow$Tr} & \textbf{En$\rightarrow$Vi} & \textbf{Avg.} \\
           \midrule
         % AT (414M) & 16.27 & 29.37 & 39.57 & 40.11 & 26.52 & 34.61 & 27.65 & 19.79 & 15.15 & 28.20 & 1$\times$ \\
         \multirow{2}{*}{\textit{Bi-lingual}} & Transformer (AT) & 432M & 1.0$\times$ & 17.0 & 30.0 & 39.8 & 39.1 & 27.2 & 34.9 & 27.0 & 19.6 & 15.0 & 28.8 & 27.8 \\
         & Transformer (NAT) & 434M & 13.4$\times$ & 6.2 & 10.6 & 25.2 & 23.0 & 14.6 & 17.5 & 13.1 & 5.3 & 0.4 & 15.2 & 13.1 \\
         % Transformer(CTC) & 173M & - & - & - & 38.2 & 37.5 & 25.8 & 32.3 & 25.4 & 18.5 & 13.2 & 27.3 & 26.2 \\
         % XLM-R-fine-tune &960M& \\
         % \midrule
         % \multirow{2}{*}{Multi-lingual} & \multicolumn{14}{l}{\textit{Multi-lingual Models}} \\
         \midrule
         \multirow{2}{*}{\textit{Multi-lingual}} & mTransformer  & - & 0.8$\times$ & 12.3 & 23.6 & 33.1 & 32.2 & 18.9 & 28.4 & 21.7 & 14.8 & 11.1 & 25.2 & 22.1 \\
         & \quad+adapter & 50M & 0.8$\times$ & 16.3 & 29.3 & 38.9 & 38.4 & 25.6 & 33.6 & 26.3 & 19.1 & 15.2 & 30.3 & 27.3 \\

         \midrule
         % \multicolumn{14}{l}{\textit{Adapt PLM}} \\
         % \midrule
         \multirow{9}{*}{\textit{PLM Adaptation}} & mGPT + MSP (AT) & 19M & 0.2$\times$ & 11.6 & 24.1 & 31.7 & 32.3 & 20.7 & \textbf{29.6} & 19.2 & \textbf{17.9} & 11.7 & 24.4 & 22.3 \\
         & \multicolumn{14}{l}{\textit{XLM-R$_{base}$}} \\
         & \quad + AT initialization & 390M & 0.9$\times$ & 7.9 & 17.6 & 26.4 & 22.6 & 14.1 & 20.1 & 15.8 & 10.6 & 6.9 & 18.4 & 16.0 \\
         & \quad + SGA w/o. denoising & 6M & 6.8$\times$ & 8.7 & 19.4 & 28.8 & 23.8 & 16.9 & 25.6 & 18.8 & 11.5 & 7.2 & 22.8 & 18.4 \\
         & \quad + SGA  & 8M & 3.0$\times$ & 11.1 & 21.7 & 32.3 & 27.5 & 18.9 & 28.5 & 21.5 & 14.7 & 8.4 & 24.3 & 20.9 \\
         & \multicolumn{14}{l}{\textit{XLM-R$_{large}$}} \\
         & \quad + AT initialization & 960M & 0.6$\times$ & 9.9 & 19.8 & 29.3 & 26.2 & 17.4 & 23.1 & 18.0 & 12.2 & 11.5 & 26.0 & 19.3 \\
         & \quad + SGA w/o. denoising & 15M & 3.7$\times$ & 11.3 & 22.2 & 33.4 & 30.6 & 19.5 & 26.5 & 22.0 & 13.1 & 9.9 & 24.7 & 21.3 \\
         & \quad + SGA & 21M & 1.9$\times$ & \textbf{13.1} & \textbf{25.2} & \textbf{37.1} & \textbf{34.3} & \textbf{21.3} & 29.2 & \textbf{24.6} & 15.5 & \textbf{11.8} & \textbf{26.9} &  \textbf{23.9} \\
         \bottomrule
    \end{tabular}}
    \caption{Results of X$\rightarrow$EN and EN$\rightarrow$X translation.  ``Param.'' represents the total number of trainable parameters. ``Speed'' represents the inference speed when batch size is 1. Scores in bold represent the best performance in the \textit{Adapt PLM} setting. Compared with the traditional fine-tuning method that directly adopts XLM-R as initialization, SGA brings large performance gains, with 8.2 BLUE on XLM-R$_{base}$ and 9.4 BLUE on XLM-R$_{large}$ on X$\rightarrow$EN, and with 4.9 BLEU on XLM-R$_{base}$ and 4.6 BLEU on XLM-R$_{large}$ on EN$\rightarrow$X, showing the effectiveness of SGA on adapting multilingual understanding models to multilingual generators.}
    \label{tab:tedtalk_experiments}
\end{table*}

\subsection{Baselines} 

We mainly compare the following baselines in our experiments.

\begin{itemize}
    \item Transformer (AT) \cite{vaswani2017attention}. We use the transformer-big setting. 
    \item Transformer (NAT) \cite{libovicky2018end}. We conduct NAT experiments on Transformer with CTC loss using the transformer-big setting.
     \item mTransformer. We train a multilingual AT Transformer with 12 encoder layers and 12 decoder layers on the TED multilingual translation datasets. Other hyperparameters are shared with Transformer-big. To report X$\rightarrow$En and En$\rightarrow$X results, we train two mTransformer models using all X$\rightarrow$En and En$\rightarrow$X data in TED, respectively.
     \item mTransformer + adapter. We use language-specific adapters \cite{bapna2019simple} on top of our trained mTransformer. We append adapters to both encoder layers and decoder layers and use a feed-forward layer dim of 1,024, which finally results in 50M extra parameters for each language pair.
      \item  mGPT + MSP \cite{tan2022msp}. mGPT + MSP introduces multi-stage prompting over a multilingual GPT model with 560M parameters. We implement this baseline following the same setting as the original paper.
      \item XLM-R w. AT initialization. Under this setting, we initialize the encoder of an autoregressive Transformer with the weights of XLM-R, and fine-tune the whole parameters. We use two variants of XLM-R: XLM-R$_{base}$ with 270M parameters, and XLM-R$_{large}$ with 550M parameters.

\end{itemize}

\subsection{Main Results}

% \subsubsection{Multilingual Translation in X$\rightarrow$En} \label{sec:main_exp}

\paragraph{SGA achieves large performance improvements over traditional initialization-based adaptation.}
Table \ref{tab:tedtalk_experiments} presents the multilingual translation experiments.  We find that initializing an autoregressive Transformer model from XLM-R only brings slight improvements by comparing with Transformer (NAT). We speculate the different nature of AR and NAR leads to performance degradation when using XLM-R as initializers. Compared with the traditional fine-tuning method that directly adopts XLM-R as initialization, SGA brings large performance gains, with 8.2 BLUE on XLM-R$_{base}$ and 9.4 BLUE on XLM-R$_{large}$ on X$\rightarrow$EN, and with 4.9 BLEU on XLM-R$_{base}$ and 4.6 BLEU on XLM-R$_{large}$ on EN$\rightarrow$X, showing the effectiveness of SGA on adapting multilingual understanding models to multilingual generators. With 21M trainable parameters, our method achieves comparable performance with bi-lingual counterparts and even better performance in several language directions (De, He, Tr). The bottom part presents translation results on the \textbf{En$\rightarrow$X} directions. %Although still lagging behind bilingual Transformers and mTransformer with adapters, our proposed XLM-R+SGA shows great potential by outperforming the previous \textit{Fixed PLM+Adaptation} methods, mGPT + MSP,  by 1.6 BLEU on average.

% Note that mGPT + MSP uses a based model of multilingual GPT with 560M parameters, which has a similar amount of parameters with XLM-R$_{large}$.

%\paragraph{Parameter efficient with good performance} \quad (i) 
%With only 8M additional parameters in all language directions and a frozen XLM-R$_{base}$ model (SGA$_{base}$+denoising), we can surpass the multilingual transformer in all language directions. (ii) With 21M trainable parameters, our method even achieves comparable performance with bi-lingual counterparts and even better performance in several language directions, including De, He and Tr.
% Note that we directly use the MLM head (which occupies a large number of parameters) produced by masked language modeling, which reveals the strong similarity between generation and NAR-like MLM pertaining. 

% We will provide a detailed analysis in Section \ref{sec:mlm_head_discuss}.

% \textbf{Prompt-Sharing enables a universal plugger for all X$\rightarrow$En directions} \quad UnlikemGPT + MSP which still needs one group of prompts for each translation direction, our proposed BIG has great potential for prompt sharing. Sharing all prompts further reduces the number of trainable parameters to a very small amount of parameters (21M for all directions) without sacrificing too much of the performance (1.7 BLEU degradation in average).

\paragraph{SGA shows better efficient inference over adaptation baselines.} Compared with the original mTransformer baseline including the adapter setting, our method achieves $1.9/0.8=2.4\times$ speedups with better performance. As a comparison, mGPT + MSP brings higher inference latency due to its multi-stage prompting and autoregressive features.
% With comparable performance or even better performance, all variants of our method achieve significant inference acceleration compared to mGPT + MSP.

\paragraph{Denoising brings large performance gains in all directions.} On both XLM-R$_{base}$ and XLM-R$_{large}$, our proposed denoising technique brings an average gain of 2.8 BLEU by increasing a very small amount of parameters. This confirms our conjecture that multilingual understanding models can be parameter-efficient language-specific denoisers due to the denoising pretraining nature of MLM.

\paragraph{XLM-R boots the performance of NAT} Existing NAT model (NAT+CTC) produces poor results in all language directions. It is because NAT generally requires an AT model to generate distillation datasets~\citep{gu2018non} which we do not provide in this paper. The NAR nature of XLM-R makes it possible to boost NAT performance. With SGA, XLM-R outperforms NAT baselines by a large margin, indicating that NAR pre-training can be further explored in future work to make multilingual understanding models better generators.

\footnotetext[1]{MSP uses different tokenization processing scripts for evaluation. To have a fair comparison, we reproduce the mGPT + MSP results in all language directions based on their \href{https://github.com/THUNLP-MT/PLM4MT}{public code}.}

\begin{figure}[t]
    \centering
    \includegraphics[width=1.0\linewidth]{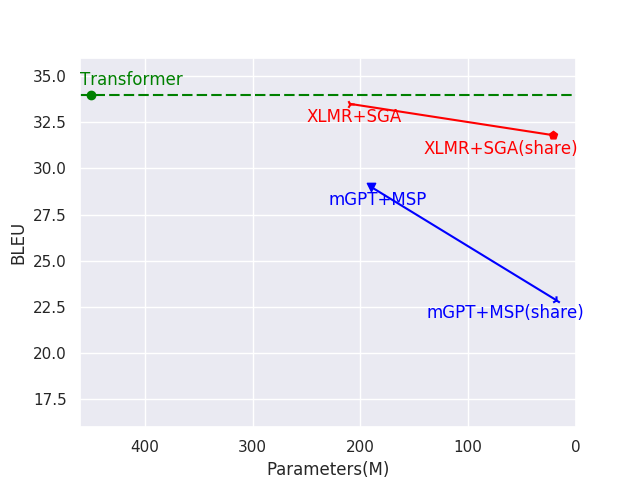}
    \caption{Tradeoff between parameter size and BLEU scores. ``share'' represents prompt sharing. Our proposed XLM-R+SGA with prompt sharing strategy can further reduce parameters without sacrificing much of the performance. As a comparison, mGPT + MSP drops significantly.}
    \label{fig:tradeoff}
\end{figure}

\begin{table*}[htp]
    \centering
    %\setlength{\tabcolsep}{9pt}
    % \fontsize{10.0pt}{\baselineskip}\selectfont
    %\resizebox{\textwidth}{!}{
    \footnotesize
    \begin{tabular}{l|lr|cccc|ccccc}
    \toprule
    \multirow{2}{*}{\textbf{Group}} & \multirow{2}{*}{\textbf{Model}} & \multirow{2}{*}{\textbf{Param.}} & \multicolumn{4}{c|}{\textbf{Supervised}} & \multicolumn{5}{c}{\textbf{Unsupervised}} \\
    % \midrule
    & & & De & Es & Fr & It & Kk & Be$^{*}$ & Eu$^{*}$ & Ms$^{*}$ & Bs$^{*}$ \\
       % \# train & 165169 & 189064 & 192756 & 201196 & 3234 & 4392 & 5094 & 5104 & 5566 & 5828 \\
    \midrule
    % mTransformer & & & & & & & & \\
    % mTransformer-all & 432M & 27.9 & 34.5 & 32.7 & 30.5 & 6.4 & 15.3 & 13.0 & 19.9 & 29.2 & 9.6 \\
    % XLM-R-fine-tune & & & & & & & & & & & \\
    % XLM-R-fine-tune & 960M & & & & & & & & & & \\
    % \midrule
    \multirow{2}{*}{\textit{Multi-lingual}} & mTransformer & 432M & 32.5 & 39.2 & 37.5 & 35.5 & 1.2 & 2.1 & 2.1 & 1.4 & 1.8 \\
    & mBART-fine-tune & 610M & \textbf{41.0} & \textbf{46.3} & \textbf{44.4} & \textbf{42.8} & \textbf{13.6} & 2.2 & 1.6 & 21.0 & 16.5 \\
    % \midrule
    % \multicolumn{11}{l}{\textit{Adapt PLM}} \\
    \midrule
    \multirow{3}{*}{\textit{Adapt PLM}}& mGPT + MSP & 19M & 27.6 & 35.2 & 33.2 & 32.0 & 6.7 & 17.1 & 10.7 & 19.3 & 14.3 \\
    & \textit{XLM-R$_{base\ }$} + SGA & 8M & 31.9 & 37.5 & 36.0 & 33.9 & 7.2 & 19.6 & 12.9 & 24.0 & 27.0 \\
    & \textit{XLM-R$_{large}$} + SGA & 21M & 34.2 & 39.3 & 38.3 & 36.2 & 11.4 & \textbf{26.8} & \textbf{20.7} & \textbf{30.1} & \textbf{34.0} \\
    \bottomrule
    \end{tabular}%}
    \caption{Zero-shot translation performance on TED in the X$\rightarrow$En directions. Our method achieves impressive performance in the zero-shot cross-lingual setting, with significant improvement in all unsupervised translation directions compared to mGPT + MSP. $^{*}$ represents that this language is not supported in mBART.}
    \label{tab:low_resource}
\end{table*}
% However, unlike circumstances in the X$\rightarrow$En directions, our method performs less well than the bi-lingual transformer model by an average of 2.3 BLEU. This motivates a further study of 

\subsection{Prompt Sharing Analysis}
We further reveal the potential of our proposed SGA by sharing prompts across languages. We combine datasets in the 10 X$\rightarrow$En language directions selected in Section \ref{sec:settings}, and compare performance with multilingual Transformer and mGPT + MSP, mBART. All baselines including mTransformer are trained using the combined dataset.

\paragraph{Prompt sharing enables a compact X$\rightarrow$En translation plugger} 
%Parameters of existing prompting methods still scale up linearly with the number of languages.
SGA achieves a better tradeoff between parameter size and inference performance by sharing prompts across all X$\rightarrow$En directions, which achieves competitive performance with the bilingual Transformer. As adaptation methods, Figure \ref{fig:tradeoff} presents the tradeoff between parameter size and performance. Performance is evaluated by averaging the test set BLEU in all directions. SGA achieves a better tradeoff performance than the AT counterpart, mGPT + MSP. With only 21M parameters, SGA enables a multilingual understanding LM a unified and impressive X$\rightarrow$En translator.

\paragraph{Prompt sharing brings impressive zero-shot cross-lingual transfer} Sharing prompts also empower SGA with strong zero-shot cross-lingual transfer ability. We choose 5 languages (Kk, Be, Eu, Ms, and Bs) with the least training data in X$\rightarrow$En directions in the TED dataset, and compare performance with multilingual Transformer, mGPT + MSP and mBART. Table \ref{tab:low_resource} presents the zero-shot cross-lingual transfer ability. (i) Trained only in 10 language directions, mTransformer outperforms mGPT + MSP in the supervised language directions with a large performance gap, while our method, XLM-R$_{large}$+SGA is still superior to both methods. (2) XLM-R+SGA achieves good performance in zero-shot X$\rightarrow$En experiments, with a substantial performance improvement when compared with mGPT + MSP and mTransformer. (3) Although still lags behind the performance of mBART on supervised language directions, mBART supports much fewer languages than XLM-R (50 vs. 100), which presents limitations in zero-shot cross-lingual performance.

\begin{table*}[tp]
    \centering
    \footnotesize
    % \small
    % \setlength{\tabcolsep}{1.8pt}
    % \fontsize{10.0pt}{\baselineskip}\selectfont
    \resizebox{\textwidth}{!}{
    \begin{tabular}{l|rr|ccc|cccccc}
    \toprule
    \multirow{3}{*}{\textbf{Models}} & \multirow{3}{*}{\textbf{Para.}} & \multirow{3}{*}{\textbf{Speed}} & \multicolumn{3}{c|}{\textbf{Question Generation}} & \multicolumn{6}{c}{\textbf{Story Generation}}  \\
    & & & En$\rightarrow$En & En$\rightarrow$De* & En$\rightarrow$Fr* & \multicolumn{2}{c}{En$\rightarrow$En} & \multicolumn{2}{c}{De$\rightarrow$En} & \multicolumn{2}{c}{Fr$\rightarrow$En} \\
    & & & RL$\uparrow$ & RL$\uparrow$ & RL$\uparrow$ & RL$\uparrow$ & Meteor$\uparrow$ & RL$\uparrow$ & Meteor$\uparrow$ & RL$\uparrow$ & Meteor$\uparrow$ \\ 
    % & BLEU & R-L & BLEU & R-L & BLEU & R-L & BLEU & R-L & BLEU & BLEU & BLEU \\
    \midrule
    % mBART fine-tune (AT) & & & 40.8 & 21.8 & 22.2 & 19.3$\dagger$ & 18.4$\dagger$ & 18.0$\dagger$ \\
    % XLM-R fine-tune (AT) & & & & & - & - & - & - \\
    % \midrule
    % \multicolumn{13}{l}{\textit{Parameter-efficient Tuning on PLMs}} \\
   mGPT + MSP & 19M & 1.0$\times$ & \textbf{36.3} & \textbf{17.5} & 17.7 & \textbf{16.7} & \textbf{16.1} & 13.7 & 13.7 & 14.2 & 13.8 \\
    % Ours$_{base}$ & 6.5 & 33.0 & 4.1 & 26.6 & 1.5 & 17.0 & 2.1 & 19.5 & 3.1 & 24.2 & 3.3 & 25.3 \\
    \midrule
   \multicolumn{12}{l}{\textit{XLM-R$_{large}$}}  \\
   \midrule
    \quad + AT initialization & 960M & 1.7$\times$ & 28.9 & 9.1 & 10.5 & 8.4 & 9.1 & 9.6 & 9.5 & 9.3 & 10.3 \\
    \quad + SGA w/o. denoising & 15M & 4.5$\times$ & 34.4 & 16.9 & 19.2 & 15.7 & 15.5 & 14.3 & 14.4 & \textbf{15.3} & 13.9 \\
    \quad + SGA & 21M & 4.3$\times$ & 35.6
    & 17.4 & \textbf{19.9} & 15.5 & 15.9 & \textbf{14.4} & \textbf{15.0} & 14.8 & \textbf{14.6} \\
    \bottomrule
    \end{tabular}}
    \caption{Results on question generation and story generation. RL represents the F1-score of Rouge-L. {*} represents zero-shot cross-lingual scenarios. SGA beats initialization-based methods on XLM-R in all cross-lingual scenarios with a substantial improvement, and achieves comparable results with mGPT + MSP.}
    \label{tab:xglue_dataset}
\end{table*}

\subsection{Other Generation Scenarios}

In this section, we test the performance of our model in various generation tasks other than multilingual translation to further explore the generation ability of multilingual understanding models.
Table \ref{tab:xglue_dataset} presents the results of both monolingual and cross-lingual results on question generation and story generation. For both tasks, we provide a monolingual result in the En$\rightarrow$En direction for reference. For the question generation task on XGLUE, since it only provides test sets in the X$\rightarrow$X directions, we train all models on the training set of En$\rightarrow$X directions, and evaluate the model performance on the X$\rightarrow$X directions for zero-shot cross-lingual generation. For the story generation task on MTG, we test supervised cross-lingual generation performance on the X$\rightarrow$En direction. We report Rouge-L scores for both tasks, and report METEOR scores additionally for story generation.

Table \ref{tab:xglue_dataset} presents the generation results. For free generation tasks, we use Mask-Predict for the alignment stage, and we set the iteration number to 4 in this table. (i) Our proposed method XLM-R+SGA can achieve comparable performance while notable acceleration, when compared with an autoregressive-based model, mGPT + MSP, on almost all generation tasks. (ii) Using XLM-R to initialize an autoregressive Transformer totally loses the zero-shot cross-lingual ability. Although it performs moderately on the supervised monolingual direction (En$\rightarrow$En) on Question Generation, it performs poorly on the zero-shot directions including En$\rightarrow$De and En$\rightarrow$Fr. (iii) Our denoising technique is proven helpful in further improving the generation quality in both tasks without sacrificing much of the speed.

% (iii) Unlike efficient tuning in understanding tasks, which have already achieved comparable performance with full-fine-tuning, efficient tuning in generation tasks still lags behind from full fine-tuning in various generation scenarios.

\begin{table}[t]
    \centering
    \small
    \begin{tabular}{p{2cm}|c|cr}
       \toprule
      \textbf{Group} & \textbf{\# Iter.}  & \textbf{Meteor$\uparrow$} & \textbf{Speed} \\
       \midrule
       \multirow{3}{*}{\textit{w/o. denoising}} &
        2 & 14.5 & 14.1 sent/s\\
        & 4 & 15.5 & 8.3 sent/s \\
       &  8 & 15.8 & 5.2 sent/s \\
        \midrule
        \multirow{4}{*}{\textit{w. denoising}} & 
        0 & 13.0 & 21.6 sent/s \\
        & 2 & 15.3 & 11.8 sent/s \\
       &  4 & 15.9 & 7.9 sent/s \\
        & 8 & 16.1 & 4.8 sent/s \\
        \bottomrule
    \end{tabular}
    \caption{Trade-off between CTC-based denoiser and number of iterations on En$\rightarrow$En generation on story generation. Batch size is set to 1. Denoising brings better performance and presents a better tradeoff between performance and inference speed.}
    \label{tab:ctc_iter_tradeoff}
\end{table}

\paragraph{Tradeoff between iterative prediction and CTC-based denoising in free generation tasks} \label{sec:tradeoff}

For free-generation tasks, unlike translation, we use iterative mask prediction instead of CTC for the alignment stage. Free generation introduces much more modalities than constrained generation tasks, specifically, translation, which intensifies the multi-modality problem in NAR generation \cite{gu2018non}. Therefore, we use an iterative method, Mask-Predict, to improve the generation quality for the alignment stage of our proposed SGA.

Although increasing the iteration number in the alignment stage can obviously lead to better performance, it will also intensify the latency problem. Our CTC-based denoiser can not only bring better performance, but also a better tradeoff between performance and speed, which is presented in Table \ref{tab:ctc_iter_tradeoff}. When the iterations of the alignment stage is set to the same, using the CTC-based denoiser leads to better performance with a slight sacrifice in speed. Using CTC with 4-step decoding can outperform 8-step decoding both in performance and speed. However, using CTC alignment alone will lead to inferior performance (0-step decoding) because of the multi-modality problem.

\section{Conclusion}

 %models have the advantage of faster generation speed and flexible decoding methods, but are only explored with limited research, especially in multilingual scenarios. 
 In this paper, we propose an effective approach to adapt existing pre-trained multilingual understanding models to multilingual generators. On translation tasks, experiments demonstrated that our proposed method achieves large performance improvements and notable acceleration with strong cross-lingual generation ability. On free-generation tasks including question generation and story generation, our method also achieves comparable performance with AT-based method with impressive speedups. Although still lagging behind pretrained multilingual AT models (e.g., mBART) in supervised fine-tuning settings in translation, our proposed method show better zero-shot abilities and faster inference. %$ the potential generation ability of multilingual understanding models with low disk requirement and faster speed. %Our findings inspire future research for better NAR pre-training to make understanding models better generators.

\section{Limitations}

Although our proposed method has achieved notable speedups and performance improvements in the multilingual setting, we still lag behind in bilingual translation, especially in high-resource scenarios. In addition, there still remains a gap between NAR pre-trained models and AR pre-trained models. Generally, XLM-R with fine-tuning is largely inferior to mBART fine-tuning. Despite the gap can be largely reduced with our method, the gap still exists. In future work, we would like to explore pre-training methods to make pretrained multilingual NAR models better generators. 
%We have conducted experiments on WMT14 English-German Dataset, which contains 4.5M translation pairs in total. In this setting, we can only achieve a similar performance with MSP, which is shown in Table \ref{tab:biling_limtate}.

% \begin{table}[h]
%     \centering
%     \begin{tabular}{c|ccc}
%     \toprule
%     Models & Transformer-big & MSP & BIG \\
%     \midrule
%     BLEU & 27.9$\dagger$ & 21.2$\dagger$ & 21.0 \\
%     \bottomrule
%     \end{tabular}
%     \caption{Bilingual Translation on WMT14 En-De dataset. $\dagger$: Results quoted from \citep{tan2022msp}.}
%     \label{tab:biling_limtate}
% \end{table}

% Like many previous works that study parameter-efficient tuning, better tuning methods are needed for enabling an understanding model for various generation tasks. Better pretraining techniques for understanding models are even needed to bridge the gap between the MLM objective and various generation tasks.


\bibliography{anthology}

\begin{thebibliography}{50}
\expandafter\ifx\csname natexlab\endcsname\relax\def\natexlab#1{#1}\fi

\bibitem[{Banerjee and Lavie(2005)}]{banerjee2005meteor}
Satanjeev Banerjee and Alon Lavie. 2005.
\newblock \href {https://aclanthology.org/W05-0909} {{METEOR}: An automatic
  metric for {MT} evaluation with improved correlation with human judgments}.
\newblock In \emph{Proceedings of the {ACL} Workshop on Intrinsic and Extrinsic
  Evaluation Measures for Machine Translation and/or Summarization}, pages
  65--72, Ann Arbor, Michigan. Association for Computational Linguistics.

\bibitem[{Bapna and Firat(2019)}]{bapna2019simple}
Ankur Bapna and Orhan Firat. 2019.
\newblock \href {https://doi.org/10.18653/v1/D19-1165} {Simple, scalable
  adaptation for neural machine translation}.
\newblock In \emph{Proceedings of the 2019 Conference on Empirical Methods in
  Natural Language Processing and the 9th International Joint Conference on
  Natural Language Processing (EMNLP-IJCNLP)}, pages 1538--1548, Hong Kong,
  China. Association for Computational Linguistics.

\bibitem[{Chen et~al.(2021)Chen, Ma, Chen, Dong, Zhang, Pan, Wang, and
  Wei}]{chen2021zero}
Guanhua Chen, Shuming Ma, Yun Chen, Li~Dong, Dongdong Zhang, Jia Pan, Wenping
  Wang, and Furu Wei. 2021.
\newblock Zero-shot cross-lingual transfer of neural machine translation with
  multilingual pretrained encoders.
\newblock In \emph{Proceedings of the 2021 Conference on Empirical Methods in
  Natural Language Processing}, pages 15--26.

\bibitem[{Chen et~al.(2022)Chen, Song, Wu, Wang, Xu, Chen, Zhou, and
  Li}]{chen2022mtg}
Yiran Chen, Zhenqiao Song, Xianze Wu, Danqing Wang, Jingjing Xu, Jiaze Chen,
  Hao Zhou, and Lei Li. 2022.
\newblock Mtg: A benchmark suite for multilingual text generation.
\newblock In \emph{Findings of the Association for Computational Linguistics:
  NAACL 2022}, pages 2508--2527.

\bibitem[{Clark et~al.(2020)Clark, Luong, Le, and Manning}]{clark2019electra}
Kevin Clark, Minh{-}Thang Luong, Quoc~V. Le, and Christopher~D. Manning. 2020.
\newblock \href {https://openreview.net/forum?id=r1xMH1BtvB} {{ELECTRA:}
  pre-training text encoders as discriminators rather than generators}.
\newblock In \emph{8th International Conference on Learning Representations,
  {ICLR} 2020, Addis Ababa, Ethiopia, April 26-30, 2020}. OpenReview.net.

\bibitem[{Conneau et~al.(2020)Conneau, Khandelwal, Goyal, Chaudhary, Wenzek,
  Guzm{\'a}n, Grave, Ott, Zettlemoyer, and Stoyanov}]{conneau2020unsupervised}
Alexis Conneau, Kartikay Khandelwal, Naman Goyal, Vishrav Chaudhary, Guillaume
  Wenzek, Francisco Guzm{\'a}n, Edouard Grave, Myle Ott, Luke Zettlemoyer, and
  Veselin Stoyanov. 2020.
\newblock \href {https://doi.org/10.18653/v1/2020.acl-main.747} {Unsupervised
  cross-lingual representation learning at scale}.
\newblock In \emph{Proceedings of the 58th Annual Meeting of the Association
  for Computational Linguistics}, pages 8440--8451, Online. Association for
  Computational Linguistics.

\bibitem[{Conneau et~al.(2018)Conneau, Rinott, Lample, Williams, Bowman,
  Schwenk, and Stoyanov}]{conneau2018xnli}
Alexis Conneau, Ruty Rinott, Guillaume Lample, Adina Williams, Samuel Bowman,
  Holger Schwenk, and Veselin Stoyanov. 2018.
\newblock \href {https://doi.org/10.18653/v1/D18-1269} {{XNLI}: Evaluating
  cross-lingual sentence representations}.
\newblock In \emph{Proceedings of the 2018 Conference on Empirical Methods in
  Natural Language Processing}, pages 2475--2485, Brussels, Belgium.
  Association for Computational Linguistics.

\bibitem[{Cooper~Stickland et~al.(2021)Cooper~Stickland, Li, and
  Ghazvininejad}]{stickland2021recipes}
Asa Cooper~Stickland, Xian Li, and Marjan Ghazvininejad. 2021.
\newblock \href {https://aclanthology.org/2021.eacl-main.301} {Recipes for
  adapting pre-trained monolingual and multilingual models to machine
  translation}.
\newblock In \emph{Proceedings of the 16th Conference of the European Chapter
  of the Association for Computational Linguistics: Main Volume}, pages
  3440--3453, Online. Association for Computational Linguistics.

\bibitem[{Devlin et~al.(2019)Devlin, Chang, Lee, and
  Toutanova}]{devlin2019bert}
Jacob Devlin, Ming-Wei Chang, Kenton Lee, and Kristina Toutanova. 2019.
\newblock \href {https://doi.org/10.18653/v1/N19-1423} {{BERT}: Pre-training of
  deep bidirectional transformers for language understanding}.
\newblock In \emph{Proceedings of the 2019 Conference of the North {A}merican
  Chapter of the Association for Computational Linguistics: Human Language
  Technologies, Volume 1 (Long and Short Papers)}, pages 4171--4186,
  Minneapolis, Minnesota. Association for Computational Linguistics.

\bibitem[{Du et~al.(2021)Du, Tu, and Jiang}]{du2021order}
Cunxiao Du, Zhaopeng Tu, and Jing Jiang. 2021.
\newblock Order-agnostic cross entropy for non-autoregressive machine
  translation.
\newblock In \emph{International Conference on Machine Learning}, pages
  2849--2859. PMLR.

\bibitem[{Fan et~al.(2021)Fan, Bhosale, Schwenk, Ma, El-Kishky, Goyal, Baines,
  Celebi, Wenzek, Chaudhary et~al.}]{fan2021beyond}
Angela Fan, Shruti Bhosale, Holger Schwenk, Zhiyi Ma, Ahmed El-Kishky,
  Siddharth Goyal, Mandeep Baines, Onur Celebi, Guillaume Wenzek, Vishrav
  Chaudhary, et~al. 2021.
\newblock Beyond english-centric multilingual machine translation.
\newblock \emph{J. Mach. Learn. Res.}, 22(107):1--48.

\bibitem[{Ghazvininejad et~al.(2020)Ghazvininejad, Karpukhin, Zettlemoyer, and
  Levy}]{ghazvininejad2020aligned}
Marjan Ghazvininejad, Vladimir Karpukhin, Luke Zettlemoyer, and Omer Levy.
  2020.
\newblock \href {http://proceedings.mlr.press/v119/ghazvininejad20a.html}
  {Aligned cross entropy for non-autoregressive machine translation}.
\newblock In \emph{Proceedings of the 37th International Conference on Machine
  Learning, {ICML} 2020, 13-18 July 2020, Virtual Event}, volume 119 of
  \emph{Proceedings of Machine Learning Research}, pages 3515--3523. {PMLR}.

\bibitem[{Ghazvininejad et~al.(2019)Ghazvininejad, Levy, Liu, and
  Zettlemoyer}]{ghazvininejad2019mask}
Marjan Ghazvininejad, Omer Levy, Yinhan Liu, and Luke Zettlemoyer. 2019.
\newblock \href {https://doi.org/10.18653/v1/D19-1633} {Mask-predict: Parallel
  decoding of conditional masked language models}.
\newblock In \emph{Proceedings of the 2019 Conference on Empirical Methods in
  Natural Language Processing and the 9th International Joint Conference on
  Natural Language Processing (EMNLP-IJCNLP)}, pages 6112--6121, Hong Kong,
  China. Association for Computational Linguistics.

\bibitem[{Gong et~al.(2022)Gong, Li, Feng, Wu, and Kong}]{gong2022diffuseq}
Shansan Gong, Mukai Li, Jiangtao Feng, Zhiyong Wu, and LingPeng Kong. 2022.
\newblock \href {https://arxiv.org/abs/2210.08933} {Diffuseq: Sequence to
  sequence text generation with diffusion models}.
\newblock \emph{ArXiv preprint}, abs/2210.08933.

\bibitem[{Gu et~al.(2018)Gu, Bradbury, Xiong, Li, and Socher}]{gu2018non}
Jiatao Gu, James Bradbury, Caiming Xiong, Victor O.~K. Li, and Richard Socher.
  2018.
\newblock \href {https://openreview.net/forum?id=B1l8BtlCb} {Non-autoregressive
  neural machine translation}.
\newblock In \emph{6th International Conference on Learning Representations,
  {ICLR} 2018, Vancouver, BC, Canada, April 30 - May 3, 2018, Conference Track
  Proceedings}. OpenReview.net.

\bibitem[{Hu et~al.(2020)Hu, Ruder, Siddhant, Neubig, Firat, and
  Johnson}]{hu2020xtreme}
Junjie Hu, Sebastian Ruder, Aditya Siddhant, Graham Neubig, Orhan Firat, and
  Melvin Johnson. 2020.
\newblock \href {http://proceedings.mlr.press/v119/hu20b.html} {{XTREME:} {A}
  massively multilingual multi-task benchmark for evaluating cross-lingual
  generalisation}.
\newblock In \emph{Proceedings of the 37th International Conference on Machine
  Learning, {ICML} 2020, 13-18 July 2020, Virtual Event}, volume 119 of
  \emph{Proceedings of Machine Learning Research}, pages 4411--4421. {PMLR}.

\bibitem[{Huang et~al.(2021)Huang, Perez, and Volkovs}]{huang2021improving}
Xiao~Shi Huang, Felipe Perez, and Maksims Volkovs. 2021.
\newblock Improving non-autoregressive translation models without distillation.
\newblock In \emph{International Conference on Learning Representations}.

\bibitem[{Kumar et~al.(2022{\natexlab{a}})Kumar, Paria, and
  Tsvetkov}]{kumar2022constrained}
Sachin Kumar, Biswajit Paria, and Yulia Tsvetkov. 2022{\natexlab{a}}.
\newblock Constrained sampling from language models via langevin dynamics in
  embedding spaces.
\newblock \emph{arXiv e-prints}, pages arXiv--2205.

\bibitem[{Kumar et~al.(2022{\natexlab{b}})Kumar, Paria, and
  Tsvetkov}]{kumar2022gradient}
Sachin Kumar, Biswajit Paria, and Yulia Tsvetkov. 2022{\natexlab{b}}.
\newblock \href {https://arxiv.org/abs/2205.12558} {Gradient-based constrained
  sampling from language models}.
\newblock \emph{ArXiv preprint}, abs/2205.12558.

\bibitem[{Lee et~al.(2018)Lee, Mansimov, and Cho}]{lee2018deterministic}
Jason Lee, Elman Mansimov, and Kyunghyun Cho. 2018.
\newblock \href {https://doi.org/10.18653/v1/D18-1149} {Deterministic
  non-autoregressive neural sequence modeling by iterative refinement}.
\newblock In \emph{Proceedings of the 2018 Conference on Empirical Methods in
  Natural Language Processing}, pages 1173--1182, Brussels, Belgium.
  Association for Computational Linguistics.

\bibitem[{Lester et~al.(2021)Lester, Al-Rfou, and Constant}]{lester2021power}
Brian Lester, Rami Al-Rfou, and Noah Constant. 2021.
\newblock The power of scale for parameter-efficient prompt tuning.
\newblock In \emph{Proceedings of the 2021 Conference on Empirical Methods in
  Natural Language Processing}, pages 3045--3059.

\bibitem[{Li and Liang(2021)}]{li2021prefix}
Xiang~Lisa Li and Percy Liang. 2021.
\newblock \href {https://doi.org/10.18653/v1/2021.acl-long.353} {Prefix-tuning:
  Optimizing continuous prompts for generation}.
\newblock In \emph{Proceedings of the 59th Annual Meeting of the Association
  for Computational Linguistics and the 11th International Joint Conference on
  Natural Language Processing (Volume 1: Long Papers)}, pages 4582--4597,
  Online. Association for Computational Linguistics.

\bibitem[{Li et~al.(2022)Li, Thickstun, Gulrajani, Liang, and
  Hashimoto}]{li2022diffusion}
Xiang~Lisa Li, John Thickstun, Ishaan Gulrajani, Percy Liang, and Tatsunori~B
  Hashimoto. 2022.
\newblock \href {https://arxiv.org/abs/2205.14217} {Diffusion-lm improves
  controllable text generation}.
\newblock \emph{ArXiv preprint}, abs/2205.14217.

\bibitem[{Liang et~al.(2020)Liang, Duan, Gong, Wu, Guo, Qi, Gong, Shou, Jiang,
  Cao, Fan, Zhang, Agrawal, Cui, Wei, Bharti, Qiao, Chen, Wu, Liu, Yang,
  Campos, Majumder, and Zhou}]{liang2020xglue}
Yaobo Liang, Nan Duan, Yeyun Gong, Ning Wu, Fenfei Guo, Weizhen Qi, Ming Gong,
  Linjun Shou, Daxin Jiang, Guihong Cao, Xiaodong Fan, Ruofei Zhang, Rahul
  Agrawal, Edward Cui, Sining Wei, Taroon Bharti, Ying Qiao, Jiun-Hung Chen,
  Winnie Wu, Shuguang Liu, Fan Yang, Daniel Campos, Rangan Majumder, and Ming
  Zhou. 2020.
\newblock \href {https://doi.org/10.18653/v1/2020.emnlp-main.484} {{XGLUE}: A
  new benchmark datasetfor cross-lingual pre-training, understanding and
  generation}.
\newblock In \emph{Proceedings of the 2020 Conference on Empirical Methods in
  Natural Language Processing (EMNLP)}, pages 6008--6018, Online. Association
  for Computational Linguistics.

\bibitem[{Libovick{\'y} and Helcl(2018)}]{libovicky2018end}
Jind{\v{r}}ich Libovick{\'y} and Jind{\v{r}}ich Helcl. 2018.
\newblock \href {https://doi.org/10.18653/v1/D18-1336} {End-to-end
  non-autoregressive neural machine translation with connectionist temporal
  classification}.
\newblock In \emph{Proceedings of the 2018 Conference on Empirical Methods in
  Natural Language Processing}, pages 3016--3021, Brussels, Belgium.
  Association for Computational Linguistics.

\bibitem[{Lin(2004)}]{lin2004rouge}
Chin-Yew Lin. 2004.
\newblock \href {https://aclanthology.org/W04-1013} {{ROUGE}: A package for
  automatic evaluation of summaries}.
\newblock In \emph{Text Summarization Branches Out}, pages 74--81, Barcelona,
  Spain. Association for Computational Linguistics.

\bibitem[{Liu et~al.(2022)Liu, Ji, Fu, Tam, Du, Yang, and Tang}]{liu2022p}
Xiao Liu, Kaixuan Ji, Yicheng Fu, Weng Tam, Zhengxiao Du, Zhilin Yang, and Jie
  Tang. 2022.
\newblock P-tuning: Prompt tuning can be comparable to fine-tuning across
  scales and tasks.
\newblock In \emph{Proceedings of the 60th Annual Meeting of the Association
  for Computational Linguistics (Volume 2: Short Papers)}, pages 61--68.

\bibitem[{Liu et~al.(2020)Liu, Gu, Goyal, Li, Edunov, Ghazvininejad, Lewis, and
  Zettlemoyer}]{liu2020multilingual}
Yinhan Liu, Jiatao Gu, Naman Goyal, Xian Li, Sergey Edunov, Marjan
  Ghazvininejad, Mike Lewis, and Luke Zettlemoyer. 2020.
\newblock \href {https://doi.org/10.1162/tacl_a_00343} {Multilingual denoising
  pre-training for neural machine translation}.
\newblock \emph{Transactions of the Association for Computational Linguistics},
  8:726--742.

\bibitem[{Liu et~al.(2019)Liu, Ott, Goyal, Du, Joshi, Chen, Levy, Lewis,
  Zettlemoyer, and Stoyanov}]{liu2019roberta}
Yinhan Liu, Myle Ott, Naman Goyal, Jingfei Du, Mandar Joshi, Danqi Chen, Omer
  Levy, Mike Lewis, Luke Zettlemoyer, and Veselin Stoyanov. 2019.
\newblock \href {https://arxiv.org/abs/1907.11692} {Roberta: A robustly
  optimized bert pretraining approach}.
\newblock \emph{ArXiv preprint}, abs/1907.11692.

\bibitem[{Ma et~al.(2021)Ma, Dong, Huang, Zhang, Muzio, Singhal,
  Hassan~Awadalla, Song, and Wei}]{ma2021deltalm}
Shuming Ma, Li~Dong, Shaohan Huang, Dongdong Zhang, Alexandre Muzio, Saksham
  Singhal, Hany Hassan~Awadalla, Xia Song, and Furu Wei. 2021.
\newblock Deltalm: Encoder-decoder pre-training for language generation and
  translation by augmenting pretrained multilingual encoders.
\newblock \emph{arXiv e-prints}, pages arXiv--2106.

\bibitem[{Mireshghallah et~al.(2022)Mireshghallah, Goyal, and
  Berg-Kirkpatrick}]{mireshghallah2022mix}
Fatemehsadat Mireshghallah, Kartik Goyal, and Taylor Berg-Kirkpatrick. 2022.
\newblock Mix and match: Learning-free controllable text generationusing energy
  language models.
\newblock In \emph{Proceedings of the 60th Annual Meeting of the Association
  for Computational Linguistics (Volume 1: Long Papers)}, pages 401--415.

\bibitem[{Ott et~al.(2019)Ott, Edunov, Baevski, Fan, Gross, Ng, Grangier, and
  Auli}]{ott2019fairseq}
Myle Ott, Sergey Edunov, Alexei Baevski, Angela Fan, Sam Gross, Nathan Ng,
  David Grangier, and Michael Auli. 2019.
\newblock fairseq: A fast, extensible toolkit for sequence modeling.
\newblock In \emph{Proceedings of the 2019 Conference of the North American
  Chapter of the Association for Computational Linguistics (Demonstrations)},
  pages 48--53.

\bibitem[{Ouyang et~al.(2021)Ouyang, Wang, Pang, Sun, Tian, Wu, and
  Wang}]{ouyang2021ernie}
Xuan Ouyang, Shuohuan Wang, Chao Pang, Yu~Sun, Hao Tian, Hua Wu, and Haifeng
  Wang. 2021.
\newblock Ernie-m: Enhanced multilingual representation by aligning
  cross-lingual semantics with monolingual corpora.
\newblock In \emph{Proceedings of the 2021 Conference on Empirical Methods in
  Natural Language Processing}, pages 27--38.

\bibitem[{Papineni et~al.(2002)Papineni, Roukos, Ward, and
  Zhu}]{papineni2002bleu}
Kishore Papineni, Salim Roukos, Todd Ward, and Wei-Jing Zhu. 2002.
\newblock \href {https://doi.org/10.3115/1073083.1073135} {{B}leu: a method for
  automatic evaluation of machine translation}.
\newblock In \emph{Proceedings of the 40th Annual Meeting of the Association
  for Computational Linguistics}, pages 311--318, Philadelphia, Pennsylvania,
  USA. Association for Computational Linguistics.

\bibitem[{Pires et~al.(2019)Pires, Schlinger, and
  Garrette}]{pires2019multilingual}
Telmo Pires, Eva Schlinger, and Dan Garrette. 2019.
\newblock \href {https://doi.org/10.18653/v1/P19-1493} {How multilingual is
  multilingual {BERT}?}
\newblock In \emph{Proceedings of the 57th Annual Meeting of the Association
  for Computational Linguistics}, pages 4996--5001, Florence, Italy.
  Association for Computational Linguistics.

\bibitem[{Post(2018)}]{post2018call}
Matt Post. 2018.
\newblock \href {https://doi.org/10.18653/v1/W18-6319} {A call for clarity in
  reporting {BLEU} scores}.
\newblock In \emph{Proceedings of the Third Conference on Machine Translation:
  Research Papers}, pages 186--191, Brussels, Belgium. Association for
  Computational Linguistics.

\bibitem[{Qi et~al.(2018)Qi, Sachan, Felix, Padmanabhan, and
  Neubig}]{qi2018and}
Ye~Qi, Devendra Sachan, Matthieu Felix, Sarguna Padmanabhan, and Graham Neubig.
  2018.
\newblock \href {https://doi.org/10.18653/v1/N18-2084} {When and why are
  pre-trained word embeddings useful for neural machine translation?}
\newblock In \emph{Proceedings of the 2018 Conference of the North {A}merican
  Chapter of the Association for Computational Linguistics: Human Language
  Technologies, Volume 2 (Short Papers)}, pages 529--535, New Orleans,
  Louisiana. Association for Computational Linguistics.

\bibitem[{Qian et~al.(2021)Qian, Zhou, Bao, Wang, Qiu, Zhang, Yu, and
  Li}]{qian2021glancing}
Lihua Qian, Hao Zhou, Yu~Bao, Mingxuan Wang, Lin Qiu, Weinan Zhang, Yong Yu,
  and Lei Li. 2021.
\newblock \href {https://doi.org/10.18653/v1/2021.acl-long.155} {Glancing
  transformer for non-autoregressive neural machine translation}.
\newblock In \emph{Proceedings of the 59th Annual Meeting of the Association
  for Computational Linguistics and the 11th International Joint Conference on
  Natural Language Processing (Volume 1: Long Papers)}, pages 1993--2003,
  Online. Association for Computational Linguistics.

\bibitem[{Qin et~al.(2022)Qin, Welleck, Khashabi, and
  Choi}]{DBLP:journals/corr/abs-2202-11705}
Lianhui Qin, Sean Welleck, Daniel Khashabi, and Yejin Choi. 2022.
\newblock {COLD} decoding: Energy-based constrained text generation with
  langevin dynamics.
\newblock \emph{CoRR}, abs/2202.11705.

\bibitem[{Saharia et~al.(2020)Saharia, Chan, Saxena, and
  Norouzi}]{saharia2020non}
Chitwan Saharia, William Chan, Saurabh Saxena, and Mohammad Norouzi. 2020.
\newblock \href {https://doi.org/10.18653/v1/2020.emnlp-main.83}
  {Non-autoregressive machine translation with latent alignments}.
\newblock In \emph{Proceedings of the 2020 Conference on Empirical Methods in
  Natural Language Processing (EMNLP)}, pages 1098--1108, Online. Association
  for Computational Linguistics.

\bibitem[{Savinov et~al.(2021)Savinov, Chung, Binkowski, Elsen, and van~den
  Oord}]{savinov2021step}
Nikolay Savinov, Junyoung Chung, Mikolaj Binkowski, Erich Elsen, and Aaron
  van~den Oord. 2021.
\newblock Step-unrolled denoising autoencoders for text generation.
\newblock In \emph{International Conference on Learning Representations}.

\bibitem[{Scao et~al.(2022)Scao, Fan, Akiki, Pavlick, Ili{\'c}, Hesslow,
  Castagn{\'e}, Luccioni, Yvon, Gall{\'e} et~al.}]{scao2022bloom}
Teven~Le Scao, Angela Fan, Christopher Akiki, Ellie Pavlick, Suzana Ili{\'c},
  Daniel Hesslow, Roman Castagn{\'e}, Alexandra~Sasha Luccioni, Fran{\c{c}}ois
  Yvon, Matthias Gall{\'e}, et~al. 2022.
\newblock \href {https://arxiv.org/abs/2211.05100} {Bloom: A 176b-parameter
  open-access multilingual language model}.
\newblock \emph{ArXiv preprint}, abs/2211.05100.

\bibitem[{Su et~al.(2021)Su, Cai, Wang, Vandyke, Baker, Li, and
  Collier}]{su2021non}
Yixuan Su, Deng Cai, Yan Wang, David Vandyke, Simon Baker, Piji Li, and Nigel
  Collier. 2021.
\newblock Non-autoregressive text generation with pre-trained language models.
\newblock In \emph{Proceedings of the 16th Conference of the European Chapter
  of the Association for Computational Linguistics: Main Volume}, pages
  234--243.

\bibitem[{Tan et~al.(2020)Tan, Zhang, Huang, Chen, Wang, Sun, Luan, and
  Liu}]{tan2020thumt}
Zhixing Tan, Jiacheng Zhang, Xuancheng Huang, Gang Chen, Shuo Wang, Maosong
  Sun, Huanbo Luan, and Yang Liu. 2020.
\newblock Thumt: an open-source toolkit for neural machine translation.
\newblock In \emph{Proceedings of the 14th Conference of the Association for
  Machine Translation in the Americas (Volume 1: Research Track)}, pages
  116--122.

\bibitem[{Tan et~al.(2022)Tan, Zhang, Wang, and Liu}]{tan2022msp}
Zhixing Tan, Xiangwen Zhang, Shuo Wang, and Yang Liu. 2022.
\newblock Msp: Multi-stage prompting for making pre-trained language models
  better translators.
\newblock In \emph{Proceedings of the 60th Annual Meeting of the Association
  for Computational Linguistics (Volume 1: Long Papers)}, pages 6131--6142.

\bibitem[{{\"U}st{\"u}n et~al.(2021){\"U}st{\"u}n, B{\'e}rard, Besacier, and
  Gall{\'e}}]{ustun2021multilingual}
Ahmet {\"U}st{\"u}n, Alexandre B{\'e}rard, Laurent Besacier, and Matthias
  Gall{\'e}. 2021.
\newblock Multilingual unsupervised neural machine translation with denoising
  adapters.
\newblock In \emph{Proceedings of the 2021 Conference on Empirical Methods in
  Natural Language Processing}, pages 6650--6662.

\bibitem[{Vaswani et~al.(2017)Vaswani, Shazeer, Parmar, Uszkoreit, Jones,
  Gomez, Kaiser, and Polosukhin}]{vaswani2017attention}
Ashish Vaswani, Noam Shazeer, Niki Parmar, Jakob Uszkoreit, Llion Jones,
  Aidan~N. Gomez, Lukasz Kaiser, and Illia Polosukhin. 2017.
\newblock \href
  {https://proceedings.neurips.cc/paper/2017/hash/3f5ee243547dee91fbd053c1c4a845aa-Abstract.html}
  {Attention is all you need}.
\newblock In \emph{Advances in Neural Information Processing Systems 30: Annual
  Conference on Neural Information Processing Systems 2017, December 4-9, 2017,
  Long Beach, CA, {USA}}, pages 5998--6008.

\bibitem[{Wang and Cho(2019)}]{DBLP:journals/corr/abs-1902-04094}
Alex Wang and Kyunghyun Cho. 2019.
\newblock \href {https://doi.org/10.18653/v1/W19-2304} {{BERT} has a mouth, and
  it must speak: {BERT} as a {M}arkov random field language model}.
\newblock In \emph{Proceedings of the Workshop on Methods for Optimizing and
  Evaluating Neural Language Generation}, pages 30--36, Minneapolis, Minnesota.
  Association for Computational Linguistics.

\bibitem[{Xu et~al.(2021)Xu, Zhou, Gan, Zheng, and Li}]{xu2021vocabulary}
Jingjing Xu, Hao Zhou, Chun Gan, Zaixiang Zheng, and Lei Li. 2021.
\newblock \href {https://doi.org/10.18653/v1/2021.acl-long.571} {Vocabulary
  learning via optimal transport for neural machine translation}.
\newblock In \emph{Proceedings of the 59th Annual Meeting of the Association
  for Computational Linguistics and the 11th International Joint Conference on
  Natural Language Processing (Volume 1: Long Papers)}, pages 7361--7373,
  Online. Association for Computational Linguistics.

\bibitem[{Xue et~al.(2021)Xue, Constant, Roberts, Kale, Al-Rfou, Siddhant,
  Barua, and Raffel}]{xue2021mt5}
Linting Xue, Noah Constant, Adam Roberts, Mihir Kale, Rami Al-Rfou, Aditya
  Siddhant, Aditya Barua, and Colin Raffel. 2021.
\newblock \href {https://doi.org/10.18653/v1/2021.naacl-main.41} {m{T}5: A
  massively multilingual pre-trained text-to-text transformer}.
\newblock In \emph{Proceedings of the 2021 Conference of the North American
  Chapter of the Association for Computational Linguistics: Human Language
  Technologies}, pages 483--498, Online. Association for Computational
  Linguistics.

\end{thebibliography}
\bibliographystyle{acl_natbib}

\clearpage
\appendix
\section{Language Code References} \label{append:lang_code}

We provide the list of languages and corresponding language codes used in our experiments in Table \ref{tab:language_code}.

\begin{table}[h]
    \centering
    \small
    \setlength{\tabcolsep}{3.2pt}
    \begin{tabular}{c|cccc}
    \toprule
     Name  & Arabic & German & Spanish & French  \\
    \midrule
     Code  & Ar & De & Es & Fr  \\
     \midrule
    \midrule
     Name  & Hebrew & Italian & Romanian & Russian  \\
     \midrule
     Code & He & It & Ro & Ru \\
     \midrule
    \midrule
     Name  &  Turkish & Vietnamese & Kazakh & Belarusian \\
    \midrule
     Code  & Tr & Vi & Kk & Be \\
    \midrule
    \midrule
    Name & Basque & Malay & Bosnian & - \\
    \midrule
     Code  & Eu & Ms & Bs & - \\
     \bottomrule
    \end{tabular}
    \caption{Full names and corresponding codes of languages used in our experiments.}
    \label{tab:language_code}
\end{table}

\section{TED Dataset Detail} \label{append:ted_detail}
Table \ref{tab:num_ted} presents a rough statistics number of the chosen 15 languages in our main experiment.
\begin{table}[h]
    \centering
    \setlength{\tabcolsep}{3pt}
    \small
    \begin{tabular}{c|rrrrr}
    \toprule
     Name  & Ar & De & Es & Fr & He \\
    \midrule
     Num.  & 211k & 165k & 193k & 189k & 208k \\
     \midrule
    \midrule
     Name  & It & Ro & Ru & Tr & Vi  \\
    \midrule
     Num.  & 201k & 178k & 205k & 180k & 169k \\
     \midrule
    \midrule
     Name  & Kk & Be & Eu & Ms & Bs \\
    \midrule
     Num.  & 3,234 & 4,392 & 5,094 & 5,104 & 5,566 \\
     \bottomrule
    \end{tabular}
    \caption{A rough statistics of the chosen 15 languages (10 for supervised setting and 5 for zero-shot cross-lingual setting) for the number of train samples in TED dataset.}
    \label{tab:num_ted}
\end{table}

% \section{Qualitative Analysis on What Denoising Brings}

% In this section, we provide a qualitative analysis of what denoising brings to the generation.

% \begin{table*}[tp]
%     \centering
%     \setlength{\tabcolsep}{12pt}
%     \fontsize{10.0pt}
%     {\baselineskip}\selectfont
%     \begin{tabular}{l|l}
%     \toprule
%     \multicolumn{2}{l}{\textit{Translation}} \\
%     \multicolumn{2}{l}{\textit{Source: Gerade werden 48 Stunden Film auf YouTube hochgeladen – pro Minute!}} \\
%     \multicolumn{2}{l}{\textit{Reference: Right now there are 48 hours of video being uploaded to YouTube every single minute.} } \\
%     \midrule
%         SGA w/o denoising & Just just 48 hours of film upload to YouTube per minute. \\
%         SGA & \textbf{Just} 48 hours of film \textbf{being uploaded} to YouTube per minute. \\
%     \midrule
%     \multicolumn{2}{l}{\textit{Question Generation}} \\
%     \multicolumn{2}{l}{\textit{Source: }} \\
%     \multicolumn{2}{l}{\textit{Reference: }} \\
%     \midrule
%         SGA w/o denoising &  \\
%         SGA & \\
%     \midrule
%     \multicolumn{2}{l}{\textit{Story Generation}} \\
%     \multicolumn{2}{l}{\textit{Source: }} \\
%     \multicolumn{2}{l}{Reference: } \\
%     \midrule
%         SGA w/o denoising &  \\
%         SGA & \\
%     \bottomrule
%     \end{tabular}
%     \caption{Qualitative Analysis on what denoising brings.}
%     \label{tab:denoise_analysis}
% \end{table*}

% From Table \ref{tab:denoise_analysis}

% \section{Example Appendix}
% \label{sec:appendix}

% This is a section in the appendix.

\end{document}